\documentclass[runningheads]{llncs}

\usepackage[T1]{fontenc}
\usepackage{graphicx}
\usepackage{amsmath}
\usepackage{amssymb}
\usepackage{subcaption}
\usepackage{url}
\usepackage{placeins}
\usepackage{todonotes}
\usepackage{float}
\usepackage{xcolor}
\usepackage{diagbox}
\usepackage{multirow}
\usepackage{authblk}
\usepackage{hyperref}
\newcommand{\discintname}{Disclosure of Interests:}

\newenvironment{credits}{\subsubsection{\discintname}\small}{}

\usepackage{etoolbox}

\makeatletter
\define@key{blx@opt@pre}{eprinttype}[]{}
\define@key{blx@opt@pre}{note}[]{}
\define@key{blx@opt@pre}{eid}[]{}
\makeatother

\usepackage[
    sorting=none,
    eprint=false, 
    eprinttype=false, 
    note=false, 
    url=false, 
    doi=false, 
    isbn=false, 
    eid=false,
    style=numeric
]{biblatex}
\begin{document}
\title{CLOFAI: A Dataset of Real And Fake Image Classification Tasks for Continual Learning}
\author{William Doherty, Anton Lee, Heitor Murilo Gomes}
\institute{Victoria University of Wellington}
%
%
%
\maketitle
\begin{abstract}
The rapid advancement of generative AI models capable of creating realistic media has led to a need for classifiers that can accurately distinguish between genuine and artificially-generated images. A significant challenge for these classifiers emerges when they encounter images from generative models that are not represented in their training data, usually resulting in diminished performance. A typical approach is to periodically update the classifier's training data with images from the new generative models then retrain the classifier on the updated dataset. However, in some real-life scenarios, storage, computational, or privacy constraints render this approach impractical. Additionally, models used in security applications may be required to rapidly adapt. In these circumstances, continual learning provides a promising alternative, as the classifier can be updated without retraining on the entire dataset. In this paper, we introduce a new dataset called CLOFAI (Continual Learning On Fake and Authentic Images), which takes the form of a domain-incremental image classification problem. Moreover, we showcase the applicability of this dataset as a benchmark for evaluating continual learning methodologies. In doing this, we set a baseline on our novel dataset using three foundational continual learning methods -- EWC, GEM, and Experience Replay -- and find that EWC performs poorly, while GEM and Experience Replay show promise, performing significantly better than a Naive baseline. The dataset and code to run the experiments can be accessed from the following GitHub repository: \url{https://github.com/Will-Doherty/CLOFAI}.

\end{abstract}
\section{Introduction}

Continual learning aims to solve the problem of catastrophic forgetting \cite{French1999}, enabling neural networks to learn new information while preserving existing knowledge. The primary challenge in continual learning is achieving an optimal balance between the ability to learn new information (plasticity) and the ability to maintain previously learned information (stability), known as the stability-plasticity dilemma \cite{DeLange2022Continual}.\\

Continual learning involves training a model on a sequence of distinct tasks, one after another. Formally, we define the data for task \( t \) as \( D_{t} = (X_{t}, Y_{t}) \), where \( X_{t} \) is the set of input data and \( Y_{t} \) is the set of corresponding output labels for task \( t \), with \( t \in T = \{1, \ldots, k\} \) for $k$ tasks. To be a continual learning problem, each task $D_{t}$ must be distinct from other tasks in either input $X_{t}$ or output $Y_{t}$ (or both). Continual learning aims to optimizes model performance across all tasks while only training on one task at a time. 
Various strategies for continual learning exist, such as regularization and replay. Regularization methods, like those proposed by \cite{kirkpatrick_overcoming_2017, zenke_continual_2017, Aljundi2018}, add explicit terms to the learning process to balance performance between old and new tasks. Replay methods, as described by \cite{lopez2017gradient}, use samples from previous tasks to supplement the training data for the current task.
\\

In this paper, we present a novel contribution to the field of continual learning by introducing CLOFAI (Continual Learning on Fake and Authentic Images), a unique dataset designed to address the challenge of detecting fake versus real images. CLOFAI is structured as a domain-incremental image classification problem, offering a diverse array of images spanning various categories and complexities. We highlight the significance of CLOFAI as a benchmark for assessing the efficacy of continual learning methodologies in the context of image classification tasks. By leveraging this dataset, researchers and practitioners can evaluate the robustness and adaptability of their continual learning algorithms in a challenging domain-incremental setting.\\

To establish a foundation for comparison, we employ three fundamental continual learning methods, setting a baseline performance on CLOFAI. Through this comparative analysis, we provide insights into the strengths and limitations of existing approaches, while discussing characteristics of the CLOFAI dataset.

\section{Related Work}

In this section, we turn to the literature on continual learning, followed by a discussion of real versus fake image classification in non-continual-learning scenarios.

\subsection{Continual Learning}

\textbf{Regularization approaches} are popular strategies to cope with forgetting, where network parameters are regularized selectively. Elastic Weight Consolidation (EWC) \cite{kirkpatrick_overcoming_2017} leverages the Fisher Information Matrix (FIM), the diagonals of which provide a parameter importance measure, to protect critical parameters during the learning of new tasks. This is done via a regularization term in the loss function that penalizes changes to parameters in proportion to their importance:

\begin{equation}
\mathcal{L}_{\text{EWC}}(\theta) = \mathcal{L}_{\text{batch}}(\theta) + \frac{\lambda}{2} \sum_j F_{jj} (\theta_j - \theta_{j, \text{old}})^2
\label{eq:ewc_loss}
\end{equation}

Here, $\theta_j$ represents the value of parameter $j$ after the most recent training batch; $\theta_{j, \text{old}}$ is previous value of parameter $j$; $F_{jj}$ is the entry at row and column $j$ in the FIM; $\mathcal{L}_{\text{batch}}(\theta)$ is the loss on the most recent training batch; $\lambda$ is a hyperparameter controlling the strength of the regularization. The loss is higher when changes to important parameters (as measured by Fisher information) are greater.\\

Like EWC, Synaptic Intelligence (SI) \cite{zenke_continual_2017} uses an importance measure to constrain parameter updates, although the specific implementation is different; for Synaptic Intelligence, parameter importance is calculated based on its contribution to the change in the loss function during training. Memory Aware Synapses (MAS) \cite{Aljundi2018} also calculates an importance measure, although it is based on the sensitivity of the predictions to parameter changes, rather than the sensitivity of the loss function.\\

The most basic \textbf{replay method} is Experience Replay, where a selection of training samples are stored in a memory buffer, then interspersed amongst the training data for subsequent tasks. The primary challenge when applying Experience Replay is deciding which samples to store in the buffer; these samples should encode maximal information about the previous tasks. For example, Mean-of-Feature sampling selects the instances that are closest to the feature mean of each class \cite{rebuffi_icarl_2017}, while Maximally Interfered Retrieval \cite{aljundi_online_2019} prioritizes the inclusion of the training samples for which the model's predictions are most adversely impacted by parameter updates. 
Gradient Episodic Memory (GEM) \cite{lopez2017gradient} is another replay approach. For a new task $t$, GEM calculates the gradient $\nabla \mathcal{L}_{t}(\theta)$ with respect to the current model parameters $\theta$, where $\mathcal{L}$ represents the loss function. It then retrieves the stored samples from previous tasks to compute the gradients for each past task. GEM aims to find a parameter update direction that minimizes the new task's loss without increasing the loss on any previous task, formalized as solving a constrained optimization problem in equation~\ref{eq:gem_optimization}. Note that $\mathcal{L}_{1:t-1}(\theta_{\text{new}}) \leq \mathcal{L}_{1:t-1}(\theta_{\text{old}})$ means the loss on each task from $1$ to $t-1$ when using the new model parameters must be lower than the loss on the same task using the old model parameters.

\begin{equation}
\min_{\theta} \mathcal{L}_{t}(\theta) \quad \text{s.t.} \quad \mathcal{L}_{1:t-1}(\theta_{\text{new}}) \leq \mathcal{L}_{1:t-1}(\theta_{\text{old}})
\label{eq:gem_optimization}
\end{equation}

\subsection{Real Versus Fake Image Classification}

Several papers investigate fake face classification \cite{Salman2022,Hamid2023,Khodabakhsh_2018}, showing that CNNs perform strongly in this domain. Other papers have examined real versus fake image classification more generally. In one example, a CNN is used to classify the GAN-generated images \cite{Baraheem2023} in a dataset created by the authors. In another \cite{bird2023cifake}, Latent Diffusion Models \cite{Rombach_2022_CVPR} are used to recreate the images from the CIFAR-10 dataset \cite{cifar10dataset} and a CNN is subsequently used to classify them.\\

It is also possible to not only classify images as real or fake, but also determine the generative model that created them. One method \cite{guarnera2023level} is to use a multi-level approach, where a CNN (ResNet \cite{He2016}) is first used to classify the image as real or fake, then a second ResNet is trained to determine whether the fake images were generated by a GAN or by Diffusion. Two more ResNets are then employed: one to differentiate between the different types of GANs, and one to differentiate between the different types of Diffusion models.\\

As indicated by the aforementioned papers, CNNs are highly suited to most real versus fake classification tasks. However, the recent emergence of Diffusion models has notably improved the quality of fake images, which are now highly realistic. As a result, binary classification using a CNN is becoming less effective \cite{wang2023dire}. There have been two recent attempts to develop non-CNN fake image detectors that can be applied to images generated by Diffusion models. The first of these is Diffusion Reconstruction Error \cite{wang2023dire}, which involves measuring the error between an input image and its reconstruction (the reconstruction is performed by a Diffusion model). The authors find that Diffusion-generated images can be reconstructed, while real images cannot, allowing them to differentiate between the two. The second approach, Diffusion Noise Feature \cite{zhang2023diffusion}, is similar. It uses an ensemble representation to estimate the noise generated during the inverse Diffusion process. A classifier can then be trained on the result.\\

While the aforementioned papers provide valuable insights into real versus fake image classification, they do not adequately address the challenges posed by the continual emergence of new generative models. To bridge this gap, we introduce the CLOFAI dataset and provide the results of several benchmark continual learning methods.

\section{CLOFAI benchmark}

\begin{figure}[ht]
    \centering
    \includegraphics[width=\linewidth]{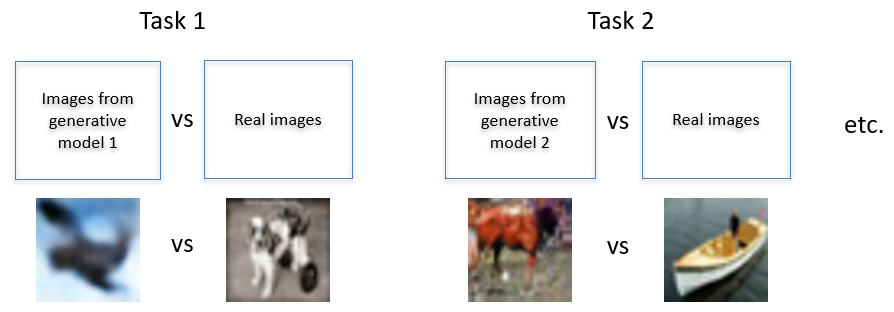}
    \caption{Problem Setup}
    \label{fig:project_overview}
\end{figure}

Figure~\ref{fig:project_overview} provides an overview of the problem setup for the CLOFAI dataset. During each task, the Classifier is trained to differentiate between real and fake images, simulating a practical scenario where a fake image detector is updated as new generative models emerge. This setup is \textit{domain-incremental} \cite{vandeven2019scenarios}, meaning the target labels are the same across tasks (in this case, the target labels are real and fake) while the input distribution changes (as different generative models are used to create the fake images for each task).\\

CLOFAI is split into five distinct tasks. For each task, there are 5000 real images, taken from the CIFAR-10 dataset \cite{cifar10dataset}, as well as 5000 fake images, created by a generative model that has been trained to reproduce the images from CIFAR-10. A different set of real images are used for each task, preventing the model from simply memorising the real images. The training and test split is 80/20, meaning there are 8000 images in the training dataset and 2000 images in the test dataset for each task. The pixel values of the images are normalised to have mean $(0, 0, 0)$ and standard deviation of $(1, 1, 1)$, as normalisation typically improves classification performance \cite{singh2020investigating}. The three elements of the tuple correspond to the three image channels - red, green, and blue.\\

\begin{figure}[ht]
    \centering
    \begin{subfigure}{0.18\textwidth}
        \centering
        \caption*{Task 1}
        \includegraphics[width=\textwidth]{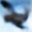}
    \end{subfigure}
    \hfill
    \begin{subfigure}{0.18\textwidth}
        \centering
        \caption*{Task 2}
        \includegraphics[width=\textwidth]{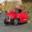}
    \end{subfigure}
    \hfill
    \begin{subfigure}{0.18\textwidth}
        \centering
        \caption*{Task 3}
        \includegraphics[width=\textwidth]{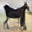}
    \end{subfigure}
    \hfill
    \begin{subfigure}{0.18\textwidth}
        \centering
        \caption*{Task 4}
        \includegraphics[width=\textwidth]{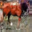}
    \end{subfigure}
    \hfill
    \begin{subfigure}{0.18\textwidth}
        \centering
        \caption*{Task 5}
        \includegraphics[width=\textwidth]{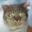}
    \end{subfigure}
    \caption{Example of a Fake Image for Each Task}
    \label{fig:image_examples}
\end{figure}

The generative models used for each task are listed below. An exemplar image from each of the models is shown in Figure~\ref{fig:image_examples}.
\begin{itemize}
    \item Task 1 -- Variational Autoencoder (VAE)~\cite{kingma2013auto} trained for this specific application.
    \item Task 2 -- A model that combines a Variational Autoencoder and an Energy-based Model~\cite{vaebm} 
    \item Task 3 -- Generative Adversarial Network (GAN)~\cite{goodfellow2014generative} trained for this specific application.
    \item Task 4 -- Flow-Based Model~\cite{densely}.
    \item Task 5 -- Denoising Diffusion Implicit Model from the HuggingFace \textit{diffusers} library~\cite{ddim_docs}.
\end{itemize}

The task order was chosen to reflect real-world circumstances where fake images become more realistic and harder to classify over time. To achieve this, we tested the Classifier's performance on each task in isolation and then ordered them such that Task 1 was the easiest and Task 5 was the hardest, with `easiest' defined as the task on which the Classifier had the highest accuracy. The results are shown in Table~\ref{table:acc_on_tasks}.\\

\begin{table}[ht]
\centering
\caption{Accuracy on Tasks}
\begin{tabular}{|l|c|}
\hline
\textbf{Task} & \textbf{Accuracy} \\ \hline
Task 1 & 99.20\% \\
Task 2 & 76.20\% \\
Task 3 & 70.65\% \\
Task 4 & 69.25\% \\
Task 5 & 63.30\% \\ \hline
\end{tabular}
\label{table:acc_on_tasks}
\end{table}

\FloatBarrier


\section{Experimental Setup}\label{sec:experiments}

The \textbf{Classifier} used in our experiments is EfficientNet\_b0 \cite{efficientnet}, with two modifications. The first modification is to add an additional linear layer which maps to a single node. The second modification is to include the sigmoid activation function after the final layer. Together, these two modifications make the model suitable for binary classification. EfficientNet\_b0 was chosen because it has been shown to have good performance on image classification benchmarks, in addition to having a relatively short training time compared to models that attain similar performance \cite{efficientnet}. In all experiments we used the weights provided in the PyTorch implementation of EfficientNet\_b0 \cite{torchvision_efficientnet_b0}, which were derived by pre-training on ImageNet \cite{deng2009imagenet}.\\

Since the classes in CLOFAI are balanced, we have used \textbf{classification accuracy} as the performance metric throughout the Experiments section. The \textbf{continual learning methods} used in the experiments are listed below: 
\begin{itemize}
    \item \textbf{Baseline:} Represents optimal performance, where the network is trained collectively on data from all encountered tasks. Note that this is not a valid continual learning approach, as continual learning, by definition, is applied in situations where not all training data is available. Instead, the baseline represents an expected upper bound on the performance of continual learning methods.
    \item \textbf{Naive:} The model is sequentially fine-tuned on each new task without strategies to prevent catastrophic forgetting, serving as a lower-bound benchmark.
    \item \textbf{Elastic Weight Consolidation (EWC)} \cite{kirkpatrick_overcoming_2017}: Assigns an importance score to each network weight, reflecting its impact on the network's performance on past tasks. Updates to weights are penalized according to their importance, with more important weights being more resistant to change. EWC serves as a representative regularization method.
    \item \textbf{Experience Replay:} Maintains a memory buffer with samples from past tasks to interleave with current task data during training, facilitating knowledge retention.
    \item \textbf{Gradient Episodic Memory (GEM)} \cite{lopez2017gradient}: Similar to Experience Replay in retaining past samples, but uses these to impose constraints on the optimization process, avoiding updates that would worsen the loss on stored instances.
\end{itemize}

We implemented each of the Baseline, Naive, EWC and Experience Replay methods from scratch. The implementation of GEM leverages the avalanche python library \cite{lomonaco2021avalanche}. The particular methods chosen were intended to encompass a range of continual learning approaches. Additionally, the continual learning community has widely adopted these methods as baselines and demonstrated them on diverse tasks~\cite{vandeven2019scenarios, DeLange2022Continual, lomonaco2021avalanche}. Note that some methods are incompatible with the (domain-incremental) problem setup, such as architecture methods that require task labels at inference. 

\section{Results}

In all experiments, the Classifier was trained on each task for 3 epochs with a batch size of 128. When the classifier is trained using more epochs it begins to overfit a particular task. The Adam optimizer \cite{kingma2014adam} was used, with a learning rate of 0.0001 and no regularization. The loss function was binary cross-entropy. All stochastic parameters had seed set to 123.

\subsection{Baseline Results}
\label{sec:base}

Table~\ref{table:baseline_pre} shows the Classifier's accuracy on the test dataset of each task. Each row in the table represents the accuracy after being trained on a particular task. Each column represents the accuracy when the Classifier is tested on a particular task. For example, the accuracy in the fourth row and second column represents the accuracy on the second task after training on the fourth task. The accuracy is coloured blue if the Classifier has already seen the task and red otherwise.

\begin{table}[ht]
\centering
\caption{Baseline Accuracy Matrix With Pre-training}
\begin{tabular}{lccccc}
\hline
\multirow{2}{*}{\diagbox[dir=NW]{\textbf{Trained}}{\textbf{Tested}}} & \multicolumn{1}{c}{\textbf{VAE}} & \multicolumn{1}{c}{\textbf{VAEBM}} & \multicolumn{1}{c}{\textbf{GAN}} & \multicolumn{1}{c}{\textbf{Flow}} & \multicolumn{1}{c}{\textbf{Diffusion}} \\
& \textbf{task1} & \textbf{task2} & \textbf{task3} & \textbf{task4} & \textbf{task5} \\ \hline
\textbf{task1} & \textcolor{blue}{99.85\%} & \textcolor{red}{49.85\%} & \textcolor{red}{50.15\%} & \textcolor{red}{50.25\%} & \textcolor{red}{50.55\%} \\
\textbf{task2} & \textcolor{blue}{94.60\%} & \textcolor{blue}{54.40\%} & \textcolor{red}{55.95\%} & \textcolor{red}{78.90\%} & \textcolor{red}{52.30\%} \\
\textbf{task3} & \textcolor{blue}{91.15\%} & \textcolor{blue}{56.30\%} & \textcolor{blue}{73.85\%} & \textcolor{red}{77.80\%} & \textcolor{red}{52.10\%} \\
\textbf{task4} & \textcolor{blue}{89.95\%} & \textcolor{blue}{67.25\%} & \textcolor{blue}{74.10\%} & \textcolor{blue}{77.60\%} & \textcolor{red}{51.60\%} \\
\textbf{task5} & \textcolor{blue}{89.15\%} & \textcolor{blue}{66.20\%} & \textcolor{blue}{73.95\%} & \textcolor{blue}{75.40\%} & \textcolor{blue}{59.05\%} \\ \hline
\end{tabular}
\label{table:baseline_pre}
\end{table}

We also tested the Classifier with random parameter initialization, i.e. no pre-training. The results, shown in Table~\ref{table:baseline_no_pre}, are far worse than the results in Table~\ref{table:baseline_pre}, indicating that pre-training provides significant benefits for classification accuracy. The primary reason that pre-training dramatically improves the Classifier's performance is due to the small number of instances in the CLOFAI training data (8,000 per task).

\begin{table}[ht]
\centering
\caption{Baseline Accuracy Matrix No Pre-Training}
\begin{tabular}{lccccc}
\hline
\multirow{2}{*}{\diagbox[dir=NW]{\textbf{Trained}}{\textbf{Tested}}} & \textbf{VAE} & \textbf{VAEBM} & \textbf{GAN} & \textbf{Flow} & \textbf{Diffusion} \\
& \textbf{task1} & \textbf{task2} & \textbf{task3} & \textbf{task4} & \textbf{task5} \\ \hline
\textbf{task1} & \textcolor{blue}{84.85\%} & \textcolor{red}{49.45\%} & \textcolor{red}{51.20\%} & \textcolor{red}{51.65\%} & \textcolor{red}{59.15\%} \\
\textbf{task2} & \textcolor{blue}{86.40\%} & \textcolor{blue}{56.40\%} & \textcolor{red}{54.65\%} & \textcolor{red}{49.00\%} & \textcolor{red}{51.55\%} \\
\textbf{task3} & \textcolor{blue}{83.90\%} & \textcolor{blue}{58.95\%} & \textcolor{blue}{58.05\%} & \textcolor{red}{48.45\%} & \textcolor{red}{47.40\%} \\
\textbf{task4} & \textcolor{blue}{81.05\%} & \textcolor{blue}{58.55\%} & \textcolor{blue}{62.50\%} & \textcolor{blue}{48.85\%} & \textcolor{red}{49.45\%} \\
\textbf{task5} & \textcolor{blue}{81.85\%} & \textcolor{blue}{56.65\%} & \textcolor{blue}{62.65\%} & \textcolor{blue}{49.65\%} & \textcolor{blue}{54.90\%} \\ \hline
\end{tabular}
\label{table:baseline_no_pre}
\end{table}

\FloatBarrier

\subsection{Naive Results}\label{sec:naive}

The Naive continual learning strategy refers to a scenario where the Classifier is sequentially fine-tuned on each new task without any method to prevent catastrophic forgetting. Table~\ref{table:naive_pre} shows that, after training on the first task, accuracy on the test dataset is almost perfect, while accuracy on unseen tasks is no better than random guessing. After training on Task 2, accuracy on the first task falls, but accuracy on the unseen tasks starts to improve, indicating that Task 2 is sufficiently similar to Task 3 and 4 that the Classifier can generalize across these tasks to a small degree.\\

\begin{table}[ht]
\centering
\caption{Naive Accuracy Matrix With Pre-training}
\begin{tabular}{lccccc}
\hline
\multirow{2}{*}{\diagbox[dir=NW]{\textbf{Trained}}{\textbf{Tested}}} & \textbf{VAE} & \textbf{VAEBM} & \textbf{GAN} & \textbf{Flow} & \textbf{Diffusion} \\
& \textbf{task1} & \textbf{task2} & \textbf{task3} & \textbf{task4} & \textbf{task5} \\ \hline
\textbf{task1} & \textcolor{blue}{99.75\%} & \textcolor{red}{50.15\%} & \textcolor{red}{50.00\%} & \textcolor{red}{50.00\%} & \textcolor{red}{50.90\%} \\
\textbf{task2} & \textcolor{blue}{85.60\%} & \textcolor{blue}{79.85\%} & \textcolor{red}{58.95\%} & \textcolor{red}{57.35\%} & \textcolor{red}{52.70\%} \\
\textbf{task3} & \textcolor{blue}{66.55\%} & \textcolor{blue}{73.65\%} & \textcolor{blue}{69.90\%} & \textcolor{red}{58.20\%} & \textcolor{red}{50.05\%} \\
\textbf{task4} & \textcolor{blue}{37.45\%} & \textcolor{blue}{64.75\%} & \textcolor{blue}{59.50\%} & \textcolor{blue}{73.35\%} & \textcolor{red}{51.85\%} \\
\textbf{task5} & \textcolor{blue}{78.25\%} & \textcolor{blue}{54.95\%} & \textcolor{blue}{51.30\%} & \textcolor{blue}{52.85\%} & \textcolor{blue}{67.85\%} \\ \hline
\end{tabular}
\label{table:naive_pre}
\end{table}

Another interesting finding is that, after training on Task 4, the accuracy on Task 1 falls to 37.45\%. One explanation is that there are shared image features across Tasks 1 and 4, but those features are labelled differently. For example, a specific pattern might be an indicator of a fake image in Task 4, while the same pattern is associated with real images in Task 1.\\

Finally, we can see that the model is using somewhat similar features to classify real and fake images in both Task 5 and Task 1, given that the accuracy on Task 1 increases after the model has been trained on Task 5. However, it is interesting that the opposite is not true; after training on Task 1 there is no improvement in accuracy in Task 5. This probably reflects the relative difficultly of the two tasks, since Task 1 does not require the model to learn complex features. Consequently, the features learnt by the model are not sufficiently complex to allow it to classify instances from Task 5.

\FloatBarrier

\subsection{EWC Results}
\label{sec:ewc}

Elastic Weight Consolidation (EWC) is a regularization method with a hyper-parameter, $\lambda$, that controls the strength of the regularization. A higher $\lambda$ means a stronger regularizing effect, meaning we would expect performance on previous tasks to degrade less, at the cost of performance on the current task. In other words, a higher $\lambda$ prioritises stability over plasticity in terms of the stability-plasticity trade-off. Table~\ref{table:ewc_lam_100000} shows the accuracy matrix of EWC when $\lambda = 100,000$. Here, EWC performs similarly to the Naive model, indicating that the regularization is failing to mitigate catastrophic forgetting.\\

\begin{table}[ht]
\centering
\caption{EWC Accuracy Matrix, $\lambda = 100,000$}
\begin{tabular}{lccccc}
\hline
\multirow{2}{*}{\diagbox[dir=NW]{\textbf{Trained}}{\textbf{Tested}}} & \textbf{VAE} & \textbf{VAEBM} & \textbf{GAN} & \textbf{Flow} & \textbf{Diffusion} \\
& \textbf{task1} & \textbf{task2} & \textbf{task3} & \textbf{task4} & \textbf{task5} \\ \hline
\textbf{task1} & \textcolor{blue}{99.75\%} & \textcolor{red}{50.15\%} & \textcolor{red}{50.00\%} & \textcolor{red}{50.00\%} & \textcolor{red}{50.90\%} \\
\textbf{task2} & \textcolor{blue}{85.80\%} & \textcolor{blue}{79.00\%} & \textcolor{red}{60.00\%} & \textcolor{red}{55.85\%} & \textcolor{red}{52.45\%} \\
\textbf{task3} & \textcolor{blue}{62.15\%} & \textcolor{blue}{73.60\%} & \textcolor{blue}{71.25\%} & \textcolor{red}{58.05\%} & \textcolor{red}{49.50\%} \\
\textbf{task4} & \textcolor{blue}{38.40\%} & \textcolor{blue}{64.10\%} & \textcolor{blue}{58.45\%} & \textcolor{blue}{73.30\%} & \textcolor{red}{52.50\%} \\
\textbf{task5} & \textcolor{blue}{75.20\%} & \textcolor{blue}{58.75\%} & \textcolor{blue}{53.85\%} & \textcolor{blue}{54.65\%} & \textcolor{blue}{66.15\%} \\ \hline
\end{tabular}
\label{table:ewc_lam_100000}
\end{table}

The natural response is to increase the value of $\lambda$, thereby increasing the strength of the regularization. However, it turns out that, irrespective of the value of $\lambda$, EWC cannot achieve good performance. Once $\lambda$ gets sufficiently high, the Classifier starts failing to learn new tasks. To illustrate this, Table~\ref{table:ewc_lam_lots} shows the result when $\lambda$ is scaled all the way to 100 million. By looking along the diagonal, it is clear that the Classifier's performance on the most recent task is weaker (relative to Table~\ref{table:ewc_lam_100000}), despite not achieving higher accuracy on prior tasks.

\begin{table}[ht]
\centering
\caption{EWC Accuracy Matrix, $\lambda = 100,000,000$}
\begin{tabular}{lccccc}
\hline
\multirow{2}{*}{\diagbox[dir=NW]{\textbf{Trained}}{\textbf{Tested}}} & \textbf{VAE} & \textbf{VAEBM} & \textbf{GAN} & \textbf{Flow} & \textbf{Diffusion} \\
& \textbf{task1} & \textbf{task2} & \textbf{task3} & \textbf{task4} & \textbf{task5} \\ \hline
\textbf{task1} & \textcolor{blue}{99.75\%} & \textcolor{red}{50.15\%} & \textcolor{red}{50.00\%} & \textcolor{red}{50.00\%} & \textcolor{red}{50.90\%} \\
\textbf{task2} & \textcolor{blue}{85.15\%} & \textcolor{blue}{68.75\%} & \textcolor{red}{55.90\%} & \textcolor{red}{51.50\%} & \textcolor{red}{57.45\%} \\
\textbf{task3} & \textcolor{blue}{76.50\%} & \textcolor{blue}{69.50\%} & \textcolor{blue}{61.30\%} & \textcolor{red}{55.90\%} & \textcolor{red}{53.40\%} \\
\textbf{task4} & \textcolor{blue}{52.65\%} & \textcolor{blue}{65.65\%} & \textcolor{blue}{56.75\%} & \textcolor{blue}{66.40\%} & \textcolor{red}{53.20\%} \\
\textbf{task5} & \textcolor{blue}{75.00\%} & \textcolor{blue}{59.40\%} & \textcolor{blue}{55.25\%} & \textcolor{blue}{55.85\%} & \textcolor{blue}{62.60\%} \\ \hline
\end{tabular}
\label{table:ewc_lam_lots}
\end{table}

To understand the reasons behind EWC's suboptimal performance, it is useful to understand its dependence on the parameter importance measure, calculated using Fisher Information. This metric assesses how changes in parameters affect the model's output distribution, thereby indicating the relative importance of parameters for accuracy on prior tasks. During the learning of new tasks, the importance measures act as constraints on the modification of weights, preserving the knowledge acquired from previous tasks. One potential explanation for EWC's poor performance is that the same parameters are highly important across all tasks. This would explain why EWC does not perform much better than the Naive method when $\lambda$ is relatively low -- the important parameters are being updated to improve accuracy on the next task, thereby diminishing accuracy on previous tasks. It would also explain why the Classifier struggles to learn when $\lambda$ is high -- critical parameters are not being adjusted due to intense regularization, while the remaining parameters fail to compensate.\\

To investigate this hypothesis, after the Classifier was trained on Task 1 we identified the ten most important parameters using Fisher Information values. We then compared their magnitudes to those observed after the Classifier was trained on Task 2. The percentage changes are shown in Table~\ref{table:percentage_change} (parameter 1 is the most important). Clearly, some of the important parameters are undergoing large changes, as indicated by the red highlight, giving credence to the hypothesis that the same parameters are highly important across tasks.

\begin{table}[ht]
\centering
\caption{Percentage Change in Important Parameters}
\begin{tabular}{lr}
\hline
\textbf{Parameter} & \textbf{Percentage Change} \\ \hline
Parameter 1 & \textcolor{red}{43.70\%} \\
Parameter 2 & 3.77\% \\
Parameter 3 & 8.46\% \\
Parameter 4 & \textcolor{red}{154.18\%} \\
Parameter 5 & \textcolor{red}{-105.23\%} \\
Parameter 6 & -7.17\% \\
Parameter 7 & \textcolor{red}{-84.26\%} \\
Parameter 8 & 5.40\% \\
Parameter 9 & 1.92\% \\
Parameter 10 & \textcolor{red}{88.52\%} \\ \hline
\end{tabular}
\label{table:percentage_change}
\end{table}

\subsection{Experience Replay Results}

Table~\ref{table:replay_100} shows the accuracy of Experience Replay with 100 samples replayed from each previously-seen task.\\

\begin{table}[ht]
\centering
\caption{Experience Replay Accuracy Matrix -- 100 Samples Replayed}
\begin{tabular}{lccccc}
\hline
\multirow{2}{*}{\diagbox[dir=NW]{\textbf{Trained}}{\textbf{Tested}}} & \textbf{VAE} & \textbf{VAEBM} & \textbf{GAN} & \textbf{Flow} & \textbf{Diffusion} \\
& \textbf{task1} & \textbf{task2} & \textbf{task3} & \textbf{task4} & \textbf{task5} \\ \hline
\textbf{task1} & \textcolor{blue}{99.75\%} & \textcolor{red}{50.15\%} & \textcolor{red}{50.00\%} & \textcolor{red}{50.00\%} & \textcolor{red}{50.90\%} \\
\textbf{task2} & \textcolor{blue}{85.95\%} & \textcolor{blue}{80.30\%} & \textcolor{red}{60.70\%} & \textcolor{red}{57.05\%} & \textcolor{red}{52.55\%} \\
\textbf{task3} & \textcolor{blue}{74.45\%} & \textcolor{blue}{72.95\%} & \textcolor{blue}{71.90\%} & \textcolor{red}{58.75\%} & \textcolor{red}{50.60\%} \\
\textbf{task4} & \textcolor{blue}{59.55\%} & \textcolor{blue}{67.15\%} & \textcolor{blue}{62.65\%} & \textcolor{blue}{74.55\%} & \textcolor{red}{51.55\%} \\
\textbf{task5} & \textcolor{blue}{79.00\%} & \textcolor{blue}{58.05\%} & \textcolor{blue}{52.40\%} & \textcolor{blue}{55.15\%} & \textcolor{blue}{66.25\%} \\ \hline
\end{tabular}
\label{table:replay_100}
\end{table}

Table~\ref{table:replay_500} shows the accuracy of Experience Replay with 500 samples replayed from each previously-seen task. Classifier accuracy with 500 samples is slightly better than accuracy with 100 samples, although the differences are small. This carries some notable implications, as a primary motivation for using continual learning is for scenarios where not all training data can be used. Particularly as the number of tasks increases, it is useful for the number of replayed samples to be minimised, assuming accuracy remains approximately constant.\\

\begin{table}[ht]
\centering
\caption{Experience Replay Accuracy Matrix -- 500 Samples Replayed}
\begin{tabular}{lccccc}
\hline
\multirow{2}{*}{\diagbox[dir=NW]{\textbf{Trained}}{\textbf{Tested}}} & \textbf{VAE} & \textbf{VAEBM} & \textbf{GAN} & \textbf{Flow} & \textbf{Diffusion} \\
& \textbf{task1} & \textbf{task2} & \textbf{task3} & \textbf{task4} & \textbf{task5} \\ \hline
\textbf{task1} & \textcolor{blue}{99.75\%} & \textcolor{red}{50.15\%} & \textcolor{red}{50.00\%} & \textcolor{red}{50.00\%} & \textcolor{red}{50.90\%} \\
\textbf{task2} & \textcolor{blue}{89.95\%} & \textcolor{blue}{79.90\%} & \textcolor{red}{60.35\%} & \textcolor{red}{56.25\%} & \textcolor{red}{53.25\%} \\
\textbf{task3} & \textcolor{blue}{82.75\%} & \textcolor{blue}{73.95\%} & \textcolor{blue}{71.80\%} & \textcolor{red}{58.05\%} & \textcolor{red}{48.95\%} \\
\textbf{task4} & \textcolor{blue}{82.20\%} & \textcolor{blue}{70.10\%} & \textcolor{blue}{63.45\%} & \textcolor{blue}{72.90\%} & \textcolor{red}{53.85\%} \\
\textbf{task5} & \textcolor{blue}{83.95\%} & \textcolor{blue}{65.50\%} & \textcolor{blue}{58.80\%} & \textcolor{blue}{59.70\%} & \textcolor{blue}{67.60\%} \\ \hline
\end{tabular}
\label{table:replay_500}
\end{table}

\FloatBarrier

\subsection{GEM Results}

Table~\ref{table:gem_100} shows the results when 100 samples are replayed (from each task) and Table~\ref{table:gem_500} shows the results when 500 samples are replayed. Unlike Experience Replay, the GEM results when 500 samples are replayed are significantly better, with the Classifier less prone to catastrophic forgetting. One caveat is that accuracy is lower on the task that has most recently been trained on when 500 samples are used. This can be seen by looking down the diagonal; for example, performance on Task 5 after training on Task 5 is better when 100 samples are used. This is not unexpected, as the model is less constrained in updating its weights when 100 samples are used, allowing it to achieve better performance on the most recent task.\\

\begin{table}[ht]
\centering
\caption{GEM Accuracy Matrix -- 100 samples Replayed}
\begin{tabular}{lccccc}
\hline
\multirow{2}{*}{\diagbox[dir=NW]{\textbf{Trained}}{\textbf{Tested}}} & \textbf{VAE} & \textbf{VAEBM} & \textbf{GAN} & \textbf{Flow} & \textbf{Diffusion} \\
& \textbf{task1} & \textbf{task2} & \textbf{task3} & \textbf{task4} & \textbf{task5} \\ \hline
\textbf{task1} & \textcolor{blue}{99.60\%} & \textcolor{red}{50.05\%} & \textcolor{red}{50.10\%} & \textcolor{red}{50.10\%} & \textcolor{red}{50.65\%} \\
\textbf{task2} & \textcolor{blue}{88.75\%} & \textcolor{blue}{78.35\%} & \textcolor{red}{58.45\%} & \textcolor{red}{54.80\%} & \textcolor{red}{51.75\%} \\
\textbf{task3} & \textcolor{blue}{77.70\%} & \textcolor{blue}{74.00\%} & \textcolor{blue}{72.25\%} & \textcolor{red}{56.70\%} & \textcolor{red}{49.70\%} \\
\textbf{task4} & \textcolor{blue}{51.70\%} & \textcolor{blue}{69.10\%} & \textcolor{blue}{65.70\%} & \textcolor{blue}{69.20\%} & \textcolor{red}{52.90\%} \\
\textbf{task5} & \textcolor{blue}{78.15\%} & \textcolor{blue}{63.50\%} & \textcolor{blue}{57.45\%} & \textcolor{blue}{58.85\%} & \textcolor{blue}{67.25\%} \\ \hline
\end{tabular}
\label{table:gem_100}
\end{table}

\begin{table}[ht]
\centering
\caption{GEM Accuracy Matrix -- 500 samples Replayed}
\begin{tabular}{lccccc}
\hline
\multirow{2}{*}{\diagbox[dir=NW]{\textbf{Trained}}{\textbf{Tested}}} & \textbf{VAE} & \textbf{VAEBM} & \textbf{GAN} & \textbf{Flow} & \textbf{Diffusion} \\
& \textbf{task1} & \textbf{task2} & \textbf{task3} & \textbf{task4} & \textbf{task5} \\ \hline
\textbf{task1} & \textcolor{blue}{99.60\%} & \textcolor{red}{50.05\%} & \textcolor{red}{50.10\%} & \textcolor{red}{50.10\%} & \textcolor{red}{50.65\%} \\
\textbf{task2} & \textcolor{blue}{91.30\%} & \textcolor{blue}{76.90\%} & \textcolor{red}{57.00\%} & \textcolor{red}{53.50\%} & \textcolor{red}{52.70\%} \\
\textbf{task3} & \textcolor{blue}{82.85\%} & \textcolor{blue}{75.45\%} & \textcolor{blue}{72.85\%} & \textcolor{red}{55.35\%} & \textcolor{red}{50.75\%} \\
\textbf{task4} & \textcolor{blue}{64.85\%} & \textcolor{blue}{72.45\%} & \textcolor{blue}{68.45\%} & \textcolor{blue}{68.55\%} & \textcolor{red}{50.90\%} \\
\textbf{task5} & \textcolor{blue}{82.25\%} & \textcolor{blue}{68.35\%} & \textcolor{blue}{63.50\%} & \textcolor{blue}{61.40\%} & \textcolor{blue}{63.00\%} \\ \hline
\end{tabular}
\label{table:gem_500}
\end{table}

The disparity in performance gains between GEM and Experience Replay when the number of replayed samples is increased from 100 to 500 can be understood through the dynamics by which the two methods operate. Experience Replay enhances learning by mixing replayed samples from previous tasks with current task data, effectively adding a term to the loss function that represents the error on the replayed samples. In this case, it appears that adding more samples simply increases the quantity of data without fundamentally changing the model's approach to balancing new learning and knowledge retention. In contrast, GEM directly constrains the optimization process by ensuring that updates to the model do not increase the loss on instances in a memory buffer. With 500 samples, GEM has a more comprehensive and varied set of constraints, derived from a wider array of past learning experiences. Evidently, this allows for a more precisely-guided optimization process.

\FloatBarrier

\section{Conclusion}

We introduced a new dataset, CLOFAI, specifically designed for real and fake image classification in a domain-incremental continual learning scenario. To establish an initial benchmark for CLOFAI, we evaluated its performance using three continual learning methods: Experience Replay, GEM, and EWC. Our findings indicate that Experience Replay and GEM demonstrated strong performance, while EWC performed poorly. Additionally, we observed a significant improvement in classifier performance with pre-training.

The continual learning approach for real and fake image classification discussed in this paper enables classifiers to efficiently adapt by training on new images without losing previously acquired knowledge. This strategy is promising for keeping pace with the rapid advancements in image generation technology. Furthermore, the CLOFAI dataset provides researchers with a valuable tool to evaluate and enhance the quality of their continual learning methods in this critical domain.

A potential direction for future research is to explore the integration of advanced generative models, such as those based on transformer architectures, into the CLOFAI dataset. Investigating how these models impact the performance and adaptability of continual learning methods could provide deeper insights and further improve classifier robustness.

\begin{credits}
The authors have no competing interests to declare that are relevant to the content of this article.
\end{credits}

\printbibliography
\end{document}